# Analyzing the Performance of Mutation Operators to Solve the Travelling Salesman Problem

Otman ABDOUN, Jaafar ABOUCHABAKA, Chakir TAJANI

LaRIT Laboratory, Faculty of sciences, Ibn Tofail University, Kenitra, Morocco
*otman.fsk@gmail.com, abouch06-univ@yahoo.fr, chakir_tajani@hotmail.fr*

**Abstract.** The genetic algorithm includes some parameters that should be adjusted, so as to get reliable results. Choosing a representation of the problem addressed, an initial population, a method of selection, a crossover operator, mutation operator, the probabilities of crossover and mutation, and the insertion method creates a variant of genetic algorithms. Our work is part of the answer to this perspective to find a solution for this combinatorial problem. What are the best parameters to select for a genetic algorithm that creates a variety efficient to solve the Travelling Salesman Problem (TSP)? In this paper, we present a comparative analysis of different mutation operators, surrounded by a dilated discussion that justifying the relevance of genetic operators chosen to solving the TSP problem.

## 1 INTRODUCTION

Nature uses several mechanisms which have led to the emergence of new species and still better adapted to their environments. The laws which react to species evolution have been known by the research of Charles Darwin in the last century[1]. We know that Darwin was the founder of the theory of evolution however John Henry Holland team's had the initiative in developing the canonical genetic algorithm, (CGA) to solve an optimization problem. Thereafter, the Goldberg work [2] is more used in a lot of applications domain. In addition to the economy, they are used in the function optimization by [3], in finance [27] and in optimal control theory in [28]-[30].

Genetic algorithms are powerful methods of optimization used successfully in different problems. Their performance is depending on the encoding scheme and the choice of genetic operators especially, the selection, crossover and mutation operators. A variety of these latest operators have been suggested in the previous researches. In particular, several operators have been developed and adapted to the permutation presentations that can be used in a large variety of combinatorial optimization problems. In this area, a typical example of the most studied problems is the Travelling Salesman Problem (TSP).

The TSP problem is classified as an NP-complete problem [24]. There are some intuitive methods to find the approximate solutions [4]-[9], but all of these methods have exponential complexity, they take too much computing time or require too much memory. In contrast to the exact algorithms, the genetic algorithms can give a good solutions but not necessary the optimal solution. These algorithms are generally very simple and have relatively a low execution time. Therefore, it may be appropriate to use a genetic algorithm [10]-[13] to solve an NP-complete problem such as the case of the TSP. These researches have provided the birth of several genetic mechanisms in particular, the selection, crossover and the mutation operators. In order to resolve the TSP problem, we propose in this paper to study empirically the impact affiliation of the selection, crossover and multiple mutation operators and finally we analyze the experimental results.

## 2 TRAVELING SALESMAN PROBLEM

The Traveling Salesman Problem (TSP) is one of the most intensively studied problems in computational mathematics. In a practical form, the problem is that a traveling salesman must visit every city in his territory exactly once and then return to the starting point [21]. Given the cost of travel between all cities, how should he plan his itinerary for minimum total cost of the entire tour?

The search space for the TSP is a set of permutations of *n* cities. Any single permutation of *n* cities yields a solution (which is a complete tour of *n* cities). The optimal solution is a permutation which yields the minimum cost of the tour. The size of the search space is *n!*

In other words, a TSP of size V is defined by a set of points v= {$v_1$, $v_2$, ..., $v_n$} which $v_i$ a city marked by coordinates $v_i.x$ and $v_i.y$ where we define a metric distance function *f* as in (1). A solution of TSP problem is a form of scheduling T=(T[1],T[2],......,T[n], T[1]) which T[i] is a permutation on the set {1, 2, ...,V}. The evaluation function calculates the adaptation of each solution of the problem by the following formula:

$$f = \sum_{i=1}^{n-1} \sqrt{(v_i.x - v_{i+1}.x)^2 + (v_i.y - v_{i+1}.y)^2} + \sqrt{(v_n.x - v_1.x)^2 + (v_n.y - v_1.y)^2} \qquad (1)$$

Where *n* is the number of cities.

If **d**, a distance matrix, is added to the TSP problem, and **d(i,j)** a distance between the city $v_i$ and $v_j$ (2), hence the cost function **f** (1) can be expressed as follows:

$$d(i,j) = \sqrt{(v_i.x - v_j.x)^2 + (v_i.y - v_j.y)^2} \quad (2)$$

$$f(T) = \sum_{i=1}^{n-1} d(T[i], T[i+1]) + d(T[n], T[1]) \quad (3)$$

The mathematical formulation of TSP problem expresses by:

$$min\{f(T), T = (T[1], T[2], \ldots\ldots, T[n])\} \quad (4)$$

Which **T[i]** is a permutation on the set {1, 2, …, V}.

The travelling salesman problem (TSP) is an *NP-hard* problem in combinatorial optimization studied in operations research and theoretical computer science [5]. A quick calculation shows that the complexity is **O(n!)** which **n** is the number of cities (Table. 1 and Figure. 1).

**Table 1.** Number of possibilities and calculation time by the number of cities

| Number of cities | Number of possibilities | Computation time |
|---|---|---|
| 5 | 12 | 12 µs |
| 10 | 181440 | 0,18 ms |
| 15 | 43 billions | 12 hours |
| 20 | 60 E+15 | 1928 years |
| 25 | 310 E+21 | 9,8 billions of years |

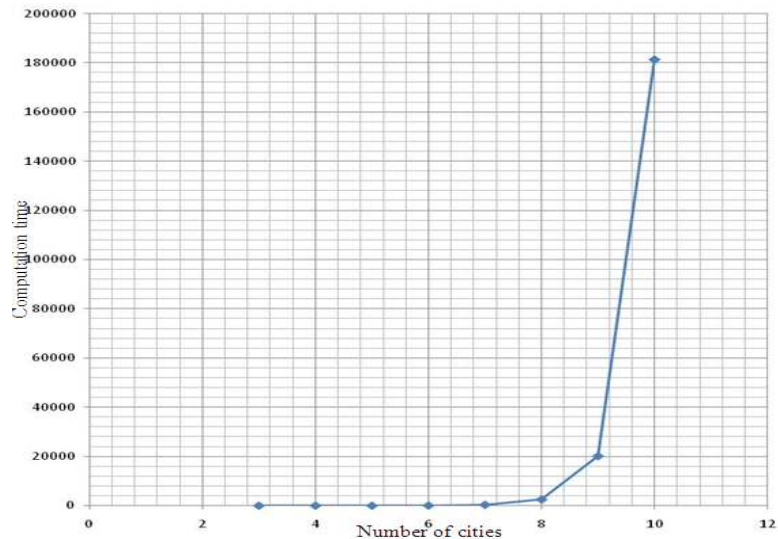

**Figure. 1.** The combinatorial explosion of TSP

To solve the TSP, there are algorithms in the literature deterministic (exact) and approximation algorithms (heuristics).

**2.1 Deterministic algorithm**

During the last decades, several algorithms emerged to approximate the optimal solution: nearest neighbor, greedy algorithm, nearest insertion, farthest insertion, double minimum spanning tree, strip, space-filling curve and Karp, Litke and Christofides algorithm, etc. (some of these algorithms assume that the cities correspond to points in the plane under some standard metric).

The TSP can be modeled in a linear programming problem under constraints, as follows:

We associate to each city a number between *1* and *V*. For each pair of cities (*i, j*), we define $c_{ij}$ the transition cost from city *i* to the city *j*, and the binary variable:

$$x_{ij} = \begin{cases} 1, & \text{If the traveler moves from city } i \text{ to city } j \\ 0, & \text{else} \end{cases} \quad (5)$$

So the TSP problem can be formulated as a problem of integer linear programming, as follows:

$$min \sum_{i=1}^{n} \sum_{j=1}^{i-1} c_{ij} \, x_{ij} \quad (6)$$

Under the following constraints:

$$1 - \sum_{i \, j} x_{ij} = 2, \forall i \in N = \{1,2,\ldots,n\} \quad (7)$$

$$2 - \sum_{i \in S} \sum_{j \notin S} x_{ij} \geq 2 \text{ for each } S \subset N \quad (8)$$

There are several deterministic algorithms; we mention the method of separation and evaluation and the method of cutting planes.

The deterministic algorithm used to find the optimal solution, but its complexity is exponential order, and it takes a lot of memory space and it requires a very high computation time. In large size problems, this algorithm cannot be used.

Because of the complexity of the problem and the limitations of the linear programming approach, other approaches are needed.

**2.2 Approximation algorithm**

An approximation algorithm, like the Genetic Algorithms, Ant Colony [31] and Tabu Search [14], [15], is a way of dealing with NP-completeness for optimization problem. This technique does not guarantee the best solution. The goal of an approximation algorithm is to come as close as possible to the optimum value in a reasonable amount of time which is at most polynomial time.

# 3  GENETIC ALGORITHM

The genetic algorithm is a one of the family of evolutionary algorithms. The population of a genetic algorithm (GA) evolves by using genetic operators inspired by the evolutionary in biology, "*The survival is the individual most suitable to the environment*". Darwin discovered that species evolution based on two components: the selection and reproduction. The selection provides a reproduction of the strongest and more robust individuals, while the reproduction is a phase in which the evolution run.

These algorithms were modeled on the natural evolution of species. We add to this evolution concepts the observed properties of genetics (Selection, Crossover, Mutation, etc), from which the name Genetic Algorithm. They attracted the interest of many researchers, starting with Holland [16], who developed the basic principles of genetic algorithm, and Goldberg [2] has used these principles to solve a specific optimization problems. Other researchers have followed this path [17]-[21].

## 3.1  Advantages

Compared to the classical optimization algorithms, the genetic algorithm has several advantages as:
- Use only the evaluation of the objective function regardless of its nature. In fact we do not require any special property of the function to be optimized (continuity, differentiability, connectedness, ..), which gives it more flexibility and a wide range of applications;
- Generation has a parallel form by working on several points at once (population of size N) instead of a single iteration in the classical algorithms;
- The use of probabilistic transition rules (crossover and mutation probability), as opposed to deterministic algorithms where the transition between two individuals is required by the structure and nature of the algorithm.

## 3.2  Principles and Functioning

Irrespective of the problems treated, genetic algorithms, presented in figure (Figure. 2), are based on six principles:
- Each treated problem has a specific way to encode the individuals of the genetic population. A chromosome (a particular solution) has different ways of being coded: numeric, symbolic, or alphanumeric;
- Creation of an initial population formed by a finite number of solutions;
- Definition of an evaluation function (fitness) to evaluate a solution;
- Selection mechanism to generate new solutions, used to identify individuals in a population that could be crossed, there are several methods in the literature, citing the method of selection by rank, roulette, by tournament, random selection, etc.;
- Reproduce the new individuals by using Genetic operators:
    1. Crossover operator: is a genetic operator that combines two chromosomes (parents) to produce a new chromosome (children) with crossover probability $P_x$ ;

2. Mutation operator: it avoids establishing a uniform population unable to evolve. This operator used to modify the genes of a chromosome selected with a mutation probability $P_m$;
- Insertion mechanism: to decide who should stay and who should disappear.
- Stopping test: to make sure about the optimality of the solution obtained by the genetic algorithm.

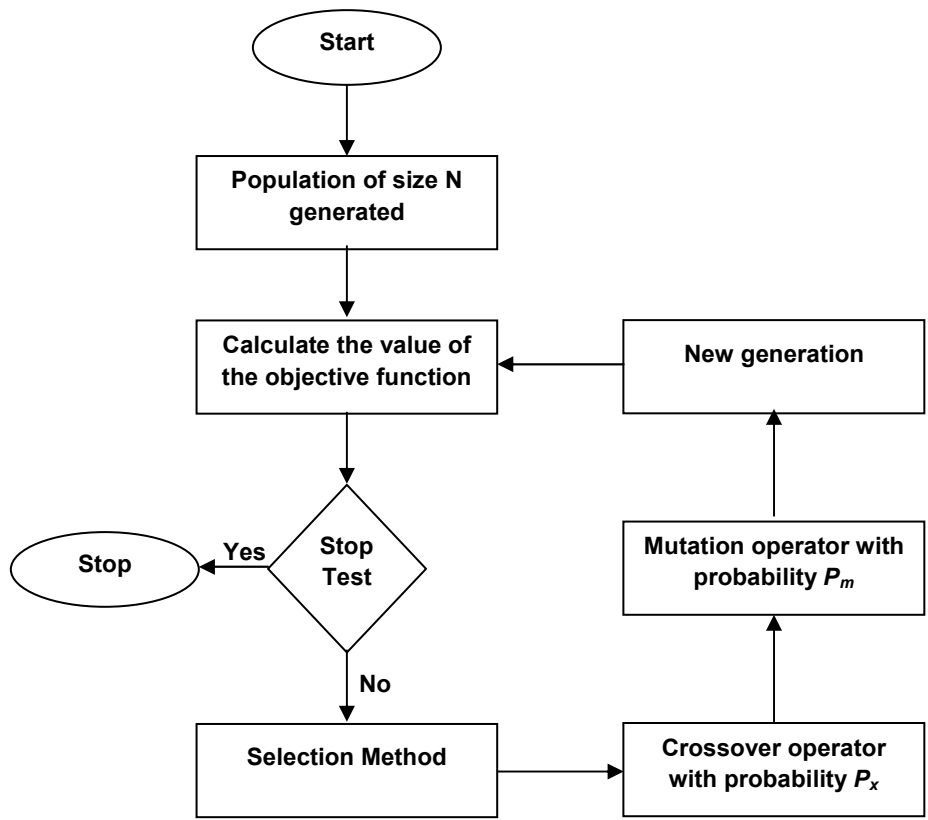

**Figure. 2.** Functioning of the genetic algorithm.

We presented the various steps which constitute the general structure of a genetic algorithm: Coding, method of selection, crossover and mutation operator and their probabilities, insertion mechanism, and the stopping test. For each of these steps, there are several possibilities. The choice between these various possibilities allows us to create several variants of genetic algorithm. Subsequently, our work focuses on finding a solution to that combinative problem: What are the best settings which create an efficient genetic variant to solve the Traveling Salesman Problem?

# 4 APPLIED GENETIC ALGORITHMS TO THE TRAVELING SALESMAN PROBLEM

## 4.1 Problem representation methods

In this section we will study different methods of data representation, and then we will quote the method used for our problem.

### 4.1.1 Adjacency Representation

The adjacency representation represents a tour as a list of *n* cities. The city *j* is listed in the position *i* if and only if the tour leads from city *i* to city *j* (Example: Table 2).

**Table 2.** Adjacency representation of the tour 1,5,2,9,7,6,8,4,1

| 5 | 2 | 9 | 7 | 6 | 8 | 4 |
|---|---|---|---|---|---|---|

The adjacency representation does not support the classical crossover operator. A repair algorithm might be necessary.

### 4.1.2 Ordinal Representation

The ordinal representation [21] represents a tour as a list of n cities; the *i th* element of the list is a number in the range from *1* to (*n-i+1*). The idea behind the ordinal representation is as follows: There is some ordered list of cities C={1,2,3,4,5,6,7,8,9}, which serves as a reference point for lists in ordinal representations.

### 4.1.3 Path Representation

The path representation is perhaps the most natural representation of a tour. A tour is encoded by an array of integers representing the successor and predecessor of each city.

**Table 3.** Coding of a tour (3, 5, 2, 9, 7, 6, 8, 4)

| 3 | 5 | 2 | 9 | 7 | 6 | 8 | 4 |
|---|---|---|---|---|---|---|---|

## 4.2 Generation of the initial population

The initial population conditions the speed and the convergence of the algorithm. For this, we applied several methods to generate the initial population:
- Random generation of the initial population.
- Generation of the first individual randomly, this one will be mutated *N-1* times with a mutation operator.

- Generation of the first individual by using a heuristic mechanism. The successor of the first city is located at a distance smaller compared to the others. Next, we use a mutation operator on the route obtained in order to generate (*N-2*) other individuals who will constitute the initial population.

**4.3 Selection**

While there are many different types of selection, we will cover the most common type - roulette wheel selection. In roulette wheel selection, the individuals are given a probability $P_i$ of being selected (9) that is directly proportionate to their fitness. The algorithm for a roulette wheel selection algorithm is illustrated in algorithm (Figure. 3)

$$\frac{1}{N-1} \times \left(1 - \frac{f_i}{\sum_{j \in Population} f_j}\right) \qquad (9)$$

Which $f_i$ is value of fitness function for the individual *i*

```
for all members of population
    sum += fitness of this individual
end for

for all members of population
    probability = sum of probabilities + (fitness / sum)
    sum of probabilities += probability
end for

number = Random between 0 and 1

for all members of population
  if number > probability but less than next probability
  then you have been selected
end for
```

**Figure. 3.** Functioning of the genetic algorithm.

Thus, individuals who have low values of the fitness function may have a high chance of being selected among the individuals to cross.

### 4.3 Crossover Operator

After the selection (reproduction) process, the population is enriched with better individuals. Reproduction makes clones of good strings but does not create new ones. Crossover operator is applied to the mating pool with the hope that it creates a better offspring.

After two parents have been selected by the selection method, crossover takes place. Crossover is an operator that mates the two parents (chromosomes) called *parent1* and *parent2* to produce two offspring (solutions) called *child1* and *child2*. The two newborn chromosomes may be better than their parents and the evolution process may continue. The crossover in carried out according to the crossover probability $P_x$. In this study, we chose as crossover operator the Ordered Crossover method (***OX)*** [33].

The Ordered Crossover method is presented by Goldberg [8], is used when the problem is of order based, for example in U-shaped assembly line balancing etc. Given two parent chromosomes, two random crossover points are selected partitioning them into a left, middle and right portion. The ordered two points crossover behaves in the following way: *child1* inherits its left and right section from *parent1*, and its middle section is determined.

**Table 4.** Crossover operator OX

|  | Parent 1 |  |  |  |  |  |  | | Parent 2 |  |  |  |  |  |  |
|---|---|---|---|---|---|---|---|---|---|---|---|---|---|---|---|
| 3 | 5 | 1 | 4 | 7 | 6 | 2 | 8 | | 4 | 7 | 5 | 1 | 8 | 6 | 2 | 3 |
| 4 | 6 | 5 | 1 | 8 | 3 | 2 | 7 | | 5 | 8 | 1 | 4 | 7 | 3 | 2 | 6 |
|  | Child 1 |  |  |  |  |  |  | | Child 2 |  |  |  |  |  |  |

We have chosen the operator OX as a crossover operator in this study because it is considered one of the best genetic operators used in the resolution of the traveling salesman problem [33].

```
Input: Parents x₁=[x_{1,1}, x_{1,2}, ……, x_{1,n}] and x₂=[x_{2,1}, x_{2,2}, ……, x_{2,n}]
Output: Children y₁=[y_{1,1}, y_{1,2}, ……, y_{1,n}] and y₂=[y_{2,1}, y_{2,2}, ……, y_{2,n}]
-------------------------------------------------------

Initialize
    • Initialize y1 and y2 being a empty genotypes;
    • Choose two crossover points a and b such that
      1≤a≤b≤n;

j₁ = j₂ = k = b+1;
i = 1;

Repeat

    if  x_{1,i} ∉ {x_{2,a}, . . . ,x_{2,b}} then
     {
       y_{1,j1} = x_{1,k} ;
       j₁++;
     }

    if  x_{2,i} ∉ {x_{1,a}, . . . ,x_{1,b}} then
     {
       y_{2,j1} = x_{2,k} ;
       j₂++;
     }

    k=k+1;
Until i ≤ n

Y₁ = [y_{1,1} ……y_{1,a-1}  x_{2,a}  ……x_{2,b}  y_{1,a}  ……y_{1,n-a}];
Y₂ = [y_{2,1} ……y_{2,a-1}  x_{1,a}  ……x_{1,b}  y_{2,a}  ……y_{2,n-a}];
```

**Figure. 4.** Algorithm of Crossover operator OX.

### 4.4 Mutation Operators

After crossover, the strings are subjected to mutation. Mutation prevents the algorithm to be trapped in a local minimum. Mutation plays the role of recovering the lost genetic materials as well as for randomly disturbing genetic information. It is an insurance policy against the irreversible loss of genetic material. Mutation has traditionally considered as a simple search operator. If crossover is supposed to exploit the current solution to find better ones, mutation is supposed to help for the exploration of the whole search space. Mutation is viewed as a background operator to maintain genetic diversity in the population. It introduces new genetic structures in the population by randomly modifying some of its building blocks. Mutation helps escape from local minima's trap and maintains diversity in the population. It also keeps the gene pool well stocked, and thus ensuring ergodicity.

There are many different forms of mutation for the different kinds of representation. For binary representation, a simple mutation can consist in inverting the value of each gene with a small probability. The probability is usually taken about 1/L, where L is the length of the chromosome. It is also possible to implement kind of hill-climbing mutation operators that do mutation only if it improves the quality of the solution. Such an operator can accelerate the search. But care should be taken, because it might also reduce the diversity in the population and makes the algorithm converge toward some local optima.

We use the sex mutation operators as following:

**4.4.1 Twors Mutation**

Twors mutation allows the exchange of position of two genes randomly chosen.

**Table 5.** Mutation operator TWORS

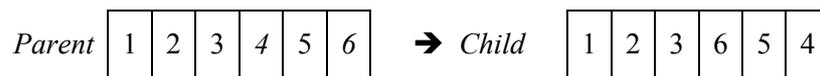

**4.4.2 Centre inverse mutation (CIM)**

The chromosome is divided into two sections. All genes in each section are copied and then inversely placed in the same section of a child.

**Table 6.** Centre inverse Mutation operator

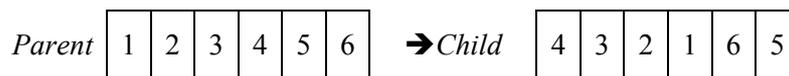

**4.4.3 Reverse Sequence Mutation (RSM)**

In the reverse sequence mutation operator, we take a sequence S limited by two positions *i* and *j* randomly chosen, such that *i<j*. The gene order in this sequence will be reversed by the same way as what has been covered in the previous operation. The algorithm (Figure. 5) shows the implementation of this mutation operator.

**Table 7.** Mutation operator RSM

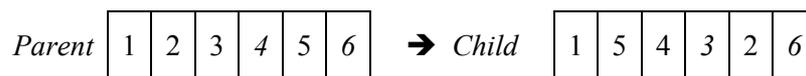

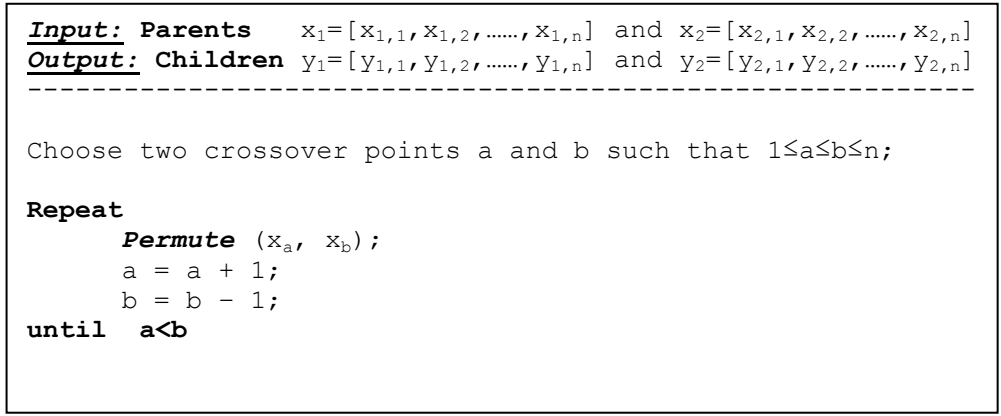

```
Input: Parents   x₁=[x_{1,1},x_{1,2},……,x_{1,n}] and x₂=[x_{2,1},x_{2,2},……,x_{2,n}]
Output: Children y₁=[y_{1,1},y_{1,2},……,y_{1,n}] and y₂=[y_{2,1},y_{2,2},……,y_{2,n}]
---------------------------------------------------------
Choose two crossover points a and b such that 1≤a≤b≤n;

Repeat
    Permute (x_a, x_b);
    a = a + 1;
    b = b - 1;
until  a<b
```

**Figure. 5.** Algorithm of RSM Operator

### 4.4.4 Throas Mutation

We construct a sequence of three genes: the first is selected randomly and the two others are those two successors. Then, the last becomes the first of the sequence, the second becomes last and the first becomes the second in the sequence.

**Table 8.** Mutation operator THORAS

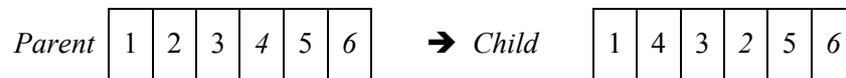

*Parent* | 1 | 2 | 3 | *4* | 5 | 6 |   ➔ *Child*   | 1 | *4* | 3 | *2* | 5 | 6 |

### 4.4.5 Thrors Mutation

Three genes are chosen randomly which shall take the different positions not necessarily successive i < j < l. the gene of the position i becomes in the position j and the one who was at this position will take the position l and the gene that has held this position takes the position i.

**Table 9.** Mutation operator RSM

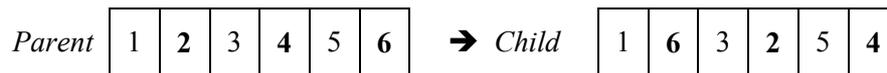

*Parent* | 1 | **2** | 3 | **4** | 5 | **6** |   ➔ *Child*   | 1 | **6** | 3 | **2** | 5 | **4** |

### 4.4.5 Partial Shuffle Mutation (PSM)

The Partial Transfer Shuffle (PSM) as its name suggests, change part of the order of the genes in the genotype. The algorithm (Figure. 6) describes in detail the stages of change.

```
Input:  Parents  x=[x₁,x₂,……,xₙ]
        and P is Mutation probability
Output: Children x=[x₁,x₂,……,xₙ]
-------------------------------------

i = 1;

Repeat

     Choose p a random number between 1 and P

     if  p < P  then
       Choose j a random number between 1 and n;
       Permute (xᵢ, xⱼ);
     End if

Until i ≤ n
```

**Figure. 2.** Algorithm of Mutation operator PSM

### 4.5 Insertion Method

We used the method of inserting *elitism* that consists in copy the best chromosome from the old to the new population [22]. This is supplemented by the solutions resulting from operations of crossover and mutation, in ensuring that the population size remains fixed from one generation to another.

We would also like to note that the GAs without elitism can also be modeled as a Markov chain proved their convergence to the limiting distributions under some conditions on the mutation probabilities [32]. However, it does not guarantee the convergence to the global optimum. With the introduction of elitism or by keeping the best string in the population allows us to show the convergence of the GA to the global optimal solution starting from any arbitrary initial population.

## 5. NUMERICAL RESULTS AND DISCUSSION

To resolve a real Traveling Salesman Problem, we use the test problem BERLIN52 to 52 locations in the city of Berlin (Figure. 7). The only optimization criterion is the distance to complete the journey. The optimal solution to this problem is known, it's 7542 m (Figure. 8).

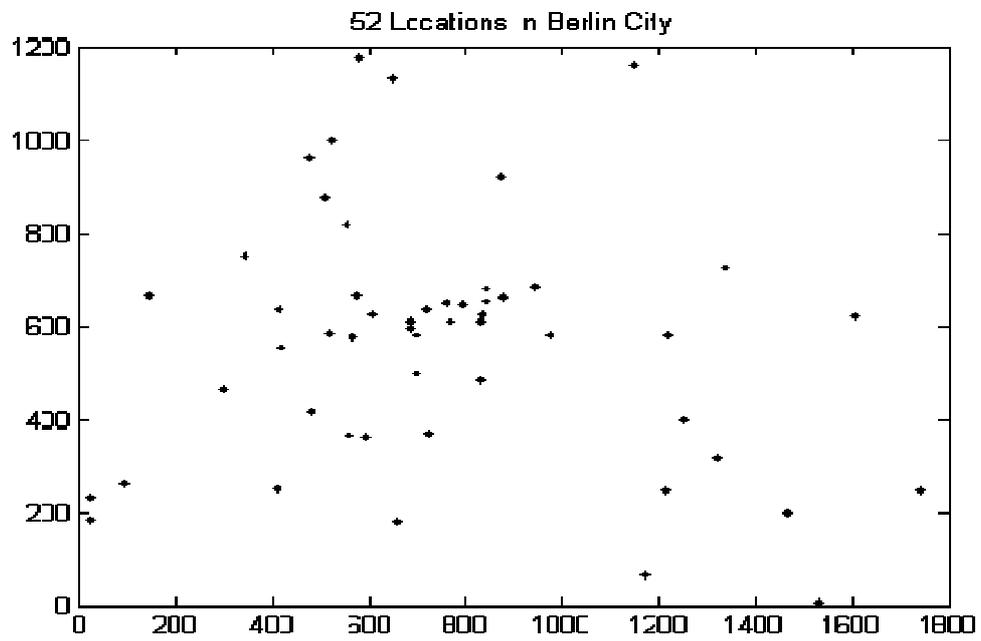

**Figure. 3.** The 52 locations in the Berlin city

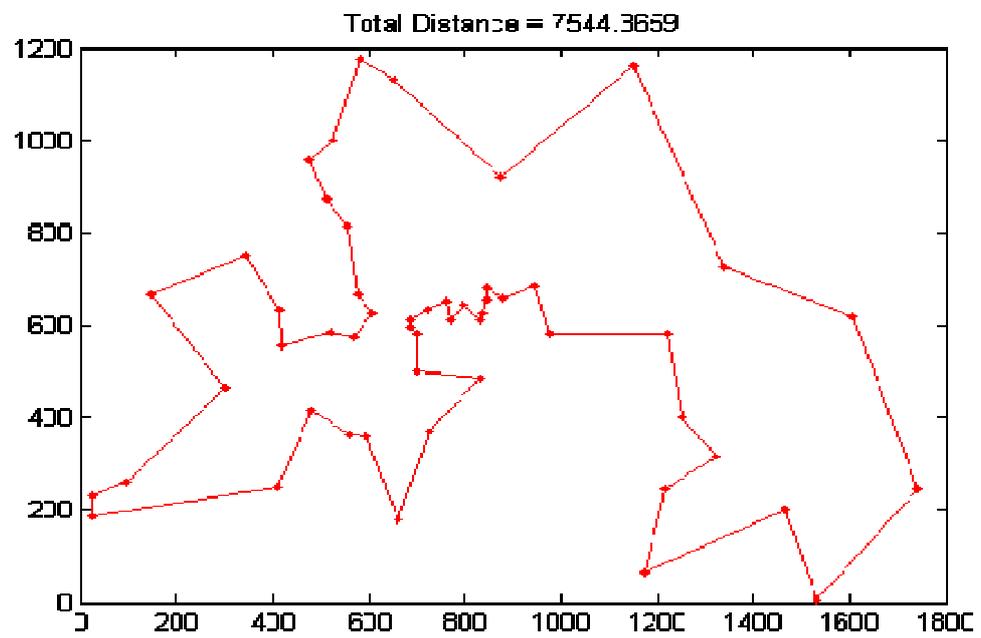

**Figure. 4.** The optimal solution of Berlin52

### 5.1 Environment

The operators of the genetic algorithm and its different modalities, which will be used later, are grouped together in the next table (table 10):

**Table 10.** The operator used

| Crossover operators | OX |
|---|---|
| Probability of crossover | 1;0.9;0.8;0.7;0.6;0.5;0.4;0.3;0.2;0.1;0 |
| Mutation operator | PSM, RSM, THRORS,THRAOS |
| Mutation probability | 1;0.9;0.8;0.7;0.6;0.5;0.4;0.3;0.2;0.1;0 |

We change at a time one parameter and we set the others and we execute the genetic algorithm fifty times. The programming was done in C++ on a PC machine with Core2Quad 2.4GHz in **CPU** and 2GB in **RAM** with a CentOS 5.5 Linux as an operating system.

### 5.2 Results and Discussion

To compare statistically the operators, these are tested one by one on 50 different initial populations after that those populations are reused for each operator. To compare the mutation operators, the algorithm used is presented in the evolutionary algorithm (Figure. 9) which the operator of variation is given by the crossover algorithm OX (Figure. 4) and followed by one of mutation operators. and the selection is made by Roulette for choosing the shortest route.

```
Generate the initial population P₀
i = 0
Repeat
        P'ᵢ = Variation (Pᵢ);
        Evaluate (P'ᵢ);
        Pᵢ₊₁ = Selection ([P'ᵢ, Pᵢ]);
Until i<Itr
```

**Figure. 5.** Evolutionary algorithm

The Figure.10 shows the results of applying a different mutation operators combined with the application of the crossover operator OX. Note that the more efficient operator is RSM, followed closely by PSM; it is interesting to note that these are two operators which cause least disturbance the individuals by moving or overturning a segment.

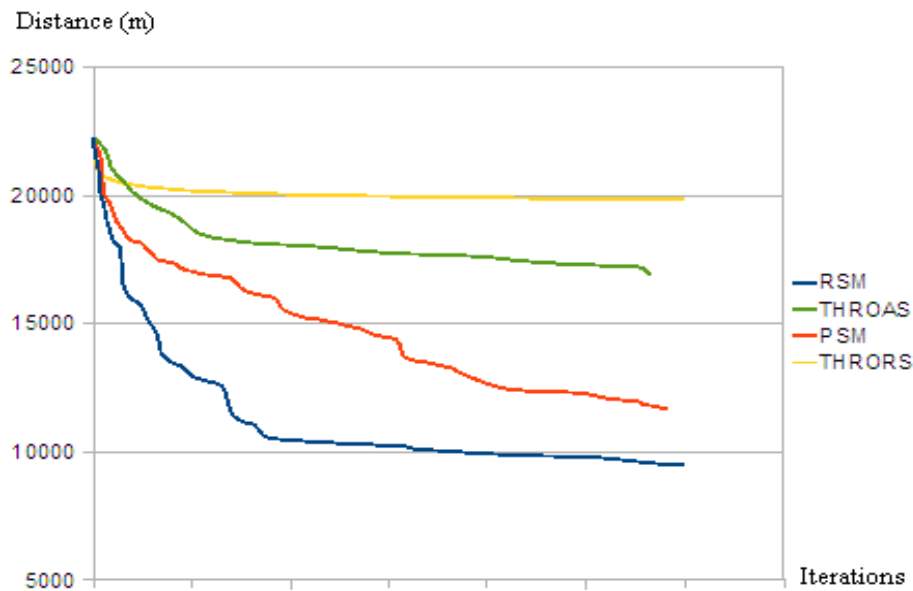

**Figure. 6.** Comparison of the mutation operators

## 6. CONCLUSION

For the traveling salesman problem presented above, the operators of mutation with the best solutions are RSM and PSM. Despite the fact that the other operators are less efficient, they can be effective with other types of problems because the solution space is different from a problem to another. The characteristics, that make an individual higher than the average, should be submitted to its descendants during the crossover and mutation steps. Base on this proposed approach (computational study), we understand why the mutation operator with a sequence moving performs better than the mixing of sequence.